# Learning Robust 3D Face Reconstruction and Discriminative Identity Representation

Yao Luo, Xiaoguang Tu, Mei Xie

University of Electronic Science and Technology of China

School of Communication & Information Engineering

Chengdu, China

luoyao_alpha@outlook.com, xguangtu@outlook.com, mxie@uestc.edu.cn

*Abstract*—3D face reconstruction from a single 2D image is a very important topic in computer vision. However, the current reconstruction methods are usually non-sensitive to face identities and over-sensitive to facial poses, which may result in similar 3D geometries for faces of different identities, or obtain different shapes for the same identity with different poses. When such methods are applied practically, their 3D estimates are either changeable for different photos of the same subject or over-regularized and generic to distinguish face identities. In this paper, we propose a robust solution to solve this problem by carefully designing a novel "Siamese" Convolutional Neural Network (SCNN). Specifically, regarding the 3D Morphable face Model (3DMM) parameters of the same individual as the same class, we employ the contrastive loss to enlarge the inter-class distance and meanwhile reduce the intra-class distance for the output 3DMM parameters. We also propose an identity loss to preserve the identity information for the same individual in the feature space. Training with these two losses, our SCNN could learn representations that are more discriminative for face identity and generalizable for pose variants. Experiments on the challenging database 300W-LP and AFLW2000-3D have shown the effectiveness of our method by comparing with state of the arts.

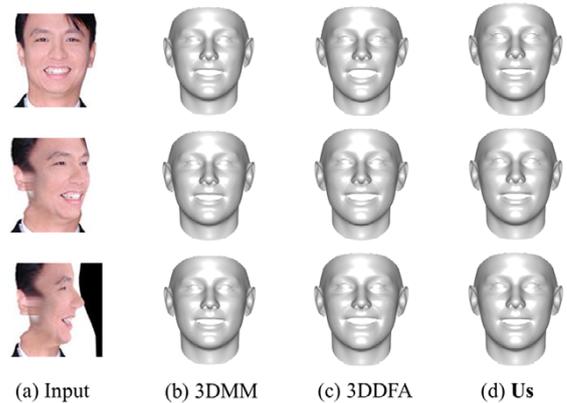

Figure 1. Different poses, 3D face shape reconstruction. (a) Input images of the same individual with different poses. (b-d) 3D reconstructions using ground truth 3DMM, (c) 3DDFA, (d) our proposed approach.

## I. INTRODUCTION

As a fundamental problem in computer vision, reconstructing 3D shapes from a single 2D face image has recently gained increasing attention, such as face alignment across large poses [5], face recognition [1] and face animation [4].

Lots of previous works have devoted to solve this problem. For instance, Zhu *et al*. propose to use end-to-end CNN based regression method with a novelty cost function to improve the face alignment performance across large poses [5], which is known as 3DDFA. Similar to [5], Liu *et al*. propose the Dense Face Alignment (DeFA) to adopt multiple constraints and multiple datasets to train a CNN that fits the 3D face to the face image [6], which achieves very dense 3D facial alignment results. In [7], Jackson *et al*. propose to map the image pixels to a volumetric representation of the 3D facial geometry through CNN-based regression, which is known as VRN. Although it bypasses the construction and fitting of a 3DMM, it requires a lot of time to predict the voxel information. In a later work [8], Feng *et al*. propose an end-to-end method called Position map Regression Network (PRN) to directly regress the complete 3D face shape along with semantic information from a single image and predict dense alignment. Although their method is effective and efficient, it is complicated to construct the UV position map. More recently, Tu *et al*. propose a novel 2D-assisted self-supervised learning (2DASL) method that can effectively use in the wild 2D face images with noisy landmark information to substantially improve 3D face model learning [9], which achieves very high performance on both 3D face reconstruction and dense face alignment.

Expected for VRN and PRN, most of the aforementioned works are based on the method that estimates the 3DMM parameters by CNN regression. However, they all face the same challenging, the recovered 3D shapes are not robust to pose variants of the same identity, and not discriminative to distinguish different subjects. It can be clearly seen from Fig. 1, our method achieves more robust results than 3DDFA, the mouths of the reconstructed shapes are different by 3DDFA, however the shapes presented by our method keep the same. The comparison results illustrate that our method is more robust than 3DDFA on reconstructing 3D face shapes from different views of faces. For face frontalization by using 3D facial shapes [17], the unstable face shapes cannot guarantee that the frontalization be well applied to different images of the same face. Therefore, the recovered 3D face shapes are rarely used for face recognition due to the unstable representation.

Overall, we observe that the popular 3DMM based methods, such as 3DDFA, DeFA and 2DASL, only consider one-to-one reconstruction process, without considering the correlation or difference within the input images, which makes the obtained results not credible for face identity representation. To address this problem, we propose a Siamese Convolutional Neural Network (SCNN) [14] which considers multi-to-one training, where the geometries of the same individual with different poses are constrained to the same, in which way the identity of the recovered shape could be preserved. The proposed method could learn more robust 3DMM parameters, and meanwhile achieve more discriminative features for identity representation.

The robustness and accuracy of our estimated shapes are verified on the 300W-LP and AFLW2000-3D dataset [18] by comparing with other 3DMM-based methods. We also show that the facial features extracted by our SCNN are more discriminative to distinguish face identities by performing experiments on the 300W-LP [18]. We will release our code and models upon the acceptance of our paper. An overview of our method is shown in Fig. 3. Our contributions are summarized as follows:

- We construct a Siamese CNN by introducing the multi-to-one strategy to train the CNN based regression methods for 3D face reconstruction.
- We make use of the identity information of the input faces and employ a novel contrastive loss to enlarge the inter-class distance and meanwhile reduce the intra-class distance for the output 3DMM parameters.
- We propose a novel identity loss to introduce the face recognition process to the 3D face reconstruction, making the learnt representations more suitable for face identity representation.

II. PROPOSED METHOD

In this part, we explain the details of the proposed method which simultaneous performing 3D face reconstruction and 2D face feature extraction. Firstly, we revisit the popular 3D morphable model (3DMM) that used to recover the 3D facial geometry [10]. Then, we elaborate on the architecture of our SCNN. Finally, we introduce the two novel loss functions that used for training.

*A. 3D Morphable Model*

Our method adopts the classical 3D morphable model (3DMM) to obtain the 3D face shape from a single face image. The 3DMM renders 3D face shape $S \in \mathbb{R}^{3N}$ that stores 3D coordinates of $N$ mesh vertices with linear combination over a set of PCA basis [9]. In [5], a 3D face shape can be formulated as follows:

$$S = \bar{S} + A_{shp} u_{shp} + A_{exp} u_{exp}. \quad (1)$$

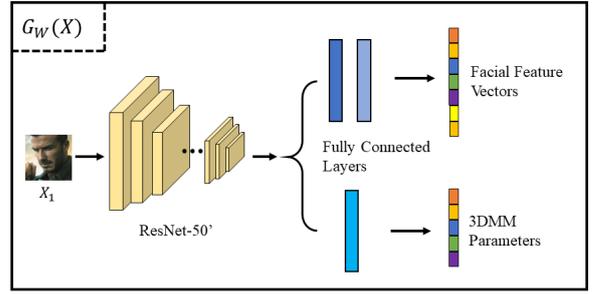

Figure 2. Network architecture

Where $\bar{S} \in \mathbb{R}^{3N}$ is the mean shape, $A_{shp} \in \mathbb{R}^{3N \times 40}$ is the shape principle components generated from the Basel Face Model (BFM) with 40 basis [11]; $A_{exp} \in \mathbb{R}^{3N \times 10}$ is the expression principle components generated from the Face Warehouse dataset with 10 basis [12]; $u_{shp} \in \mathbb{R}^{40}$ is a shape parameter vectors and $u_{exp} \in \mathbb{R}^{10}$ is an expression parameter vectors. The 3D face is then projected onto the image plane with weak perspective projection to generate a 2D face from specified viewpoint

$$V(g) = f \times \text{Pr} \times R \times S + t_{2d}. \quad (2)$$

Where $V(g)$ stores the 2D coordinates of the 3D vertices projected onto the 2D plane, $f$ is the scale factor, control the 3D face size, Pr is the orthographic projection matrix $\begin{pmatrix} 1 & 0 & 0 \\ 0 & 1 & 0 \end{pmatrix}$, $R$ is the rotation matrix consisting of 9 parameters and $t_{2d}$ is a 2D translation vector. For ease of description, we put them together, so the collection of all the model parameters is $g = [f, R, t_{2d}, u_{shp}, u_{exp}]$, a total of 62 parameters to regress for the 3D face regressor model.

*B. Architecture of Siamese Convolutional Neural Network*

The architecture of our SCNN is given in Fig. 2. We use Resnet-50 [13] as the backbone of our SCNN, with the last two layers replaced by two parallel fully connected (FC) layers. One FC layer is a 62-dimensional vector to represent the 3DMM parameters, and the other is a 512-dimensional vector for face recognition.

The general framework of our SCNN is given in Fig. 3 As shown, $X_1$ and $X_2$ are the input images to our model; $W$ is the shared parameter vector that is subject to learning; $G_W(X_1)_{3d}$ and $G_W(X_2)_{3d}$ are the 3DMM parameters of the regression. Here we select 50 elements from each of $G_W(X_1)_{3d}$ and $G_W(X_2)_{3d}$ as the shape parameter $G_W(X_1)_{shp}$ and $G_W(X_2)_{shp}$ respectively. Additionally, $G_W(X_1)_{id}$ and $G_W(X_2)_{id}$ are the two representations in the feature space for the input face images respectively [15]. The whole model is trained by minimizing the following three losses.

The first one is the weighted coefficient prediction loss $\ell_{3d}$ over the 3D annotated image that measures how accurate the model can predict 3DMM coefficients [9]; the second one is the constraint loss $\ell_{shp}$ that force the reconstructed face shape be the same ignore the various poses; the third one is the identity loss $\ell_{id}$ that ensures the face images regarding the same identity have the similar distribution in the feature space. Therefore, the overall training loss is

$$\ell = \ell_{3d} + \ell_{shp} + \ell_{id}. \qquad (3)$$

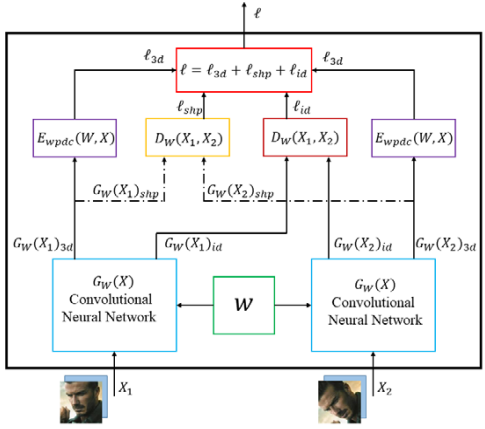

Figure 3. Siamese architecture details

### C. Loss Functions

#### 1) Weighted Parameter Distance Cost (WPDC)

Following [5] and [9], we use the weight parameter distance cost (WPDC) to train our model. According to the importance of each parameter in the estimated 3DMM parameters, we derive the following formula:

$$\ell_{3d} = \sum_{j=1}^{P} [E_{wpdc}(W,(X_1)^j) + E_{wpdc}(W,(X_2)^j)], \qquad (4)$$

$$E_{wpdc}(W,X) = (a - \hat{a})Q(a - \hat{a})^T, \qquad (5)$$

where,

$$Q = diag(q_1, q_2, ..., q_{62}), \qquad (6)$$

$$q_i = \frac{1}{\sum_i q_i} \|V(a) - V(\hat{a}_i)\|. \qquad (7)$$

Where $X_1$ and $X_2$ are the $j$-th input image-pair, $P$ is the number of training pairs, $Q$ is the importance matrix whose diagonal element indicates the importance of each parameter, $X$ is the image that belongs to one of a pair of input pictures. $\hat{a}_i$ is the coefficient vector whose $i$-th element comes from the predicted parameter and the others come from the ground truth parameter $a$, and $V(\cdot)$ is the sparse landmark projection from rendered 3D shape. In the training process, CNN firstly focuses on learning the coefficients with larger weight such as rotation and translation. After their error decreases, the CNN model turn to optimize less important parameters (e.g., the ones for shape and expression), meanwhile, maintain the high-priority coefficients sufficiently good.

#### 2) Contrastive Cost Function

Following [2], we define the parameterized distance function $D_W$ as the Euclidean distance between the outputs of $G_W(X_1)_{shp}$ and $G_W(X_2)_{shp}$. That is,

$$D_W(X_1, X_2) = \|G_W(X_1)_{shp} - G_W(X_2)_{shp}\|_2. \qquad (8)$$

For short notation, we write $D_W(X_1, X_2)$ as $D_W$. Then the loss function is described as follows:

$$\ell_{shp} = \sum_{j=1}^{P} L(W, Y, (X_1, X_2)^j), \qquad (9)$$

where,

$$L(W, Y, X_1, X_2) = (Y)\frac{1}{2}(D_W)^2 + (1-Y)\{\max(0, m - D_W)\}^2. \qquad (10)$$

Where $(Y,(X_1,X_2)^j)$ is the $j$-th labeled sample pair, $X_1$ and $X_2$ are the input image-pair, $Y=1$ if $X_1$ and $X_2$ belong to the same person (a "genuine pair") and $Y=0$ otherwise (an "impostor pair") [2], $P$ is the number of training pairs, and $m > 0$ is a constant that can be interpreted as a margin.

As for the outputs $G_W(X_1)_{id}$ and $G_W(X_2)_{id}$, we adopt the same cost function with (10), the only difference is $D_W(X_1, X_2)$, which is defined as follows:

$$D_W(X_1, X_2) = \|G_W(X_1)_{id} - G_W(X_2)_{id}\|_2. \qquad (11)$$

### III. EXPERIMENT

In this section, we test our proposed method, compare the accuracy and robustness of its estimated 3D shapes with other methods. We also evaluate the effectiveness of the extracted identity features for recognition.

### A. Training Details and Datasets

We implement our proposed method on Pytorch [19], and employ SGD as the optimizer, the learning rate decays exponentially for the CNN regressor. The learning rates of training loss function ($\ell_{3d}$, $\ell_{shp}$ $\ell_{exp}$) begin at $l_{3d}=1\times10^{-2}$, $l_{shp}=1\times10^{-3}$, $l_{id}=1\times10^{-4}$ respectively. The batch size is set as 32. In training process, we adopt a two-stage strategy to train our model. In the first stage, we train the model using the loss $\ell_{3d}$. In the second stage, we refine our model using the overall loss $\ell$.

The dataset 300W-LP [18] contains more than 60K face images with annotated 3DMM coefficients, containing overall

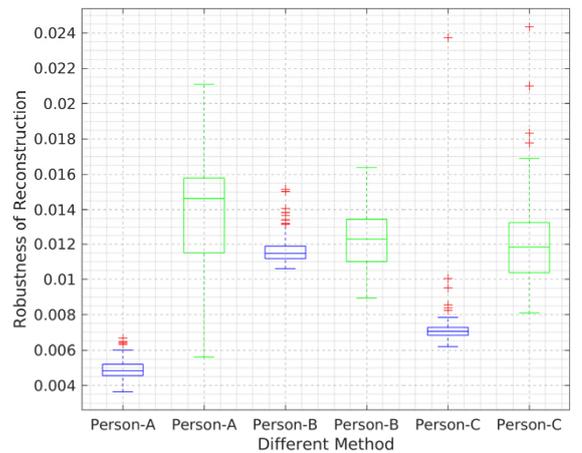

Figure 4. Error distribution of 3D face reconstruction results on SET1 dataset. Person-x represent the individual, the blue boxplots represent the results of SCNN, and others (green boxplots) represent the results of 3DDFA.

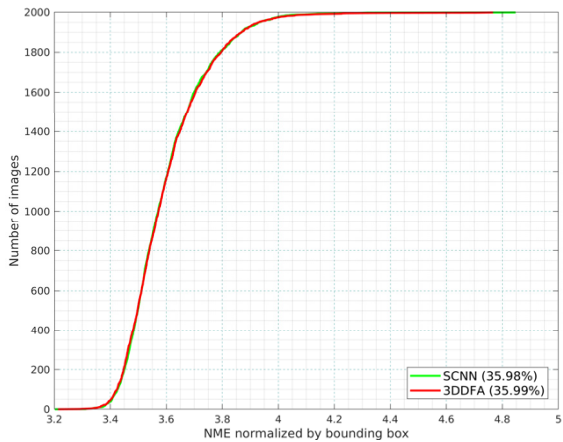

Figure 5. Error Distribution Curves (EDC) of 3D face reconstruction results on AFLW2000-3D dataset.

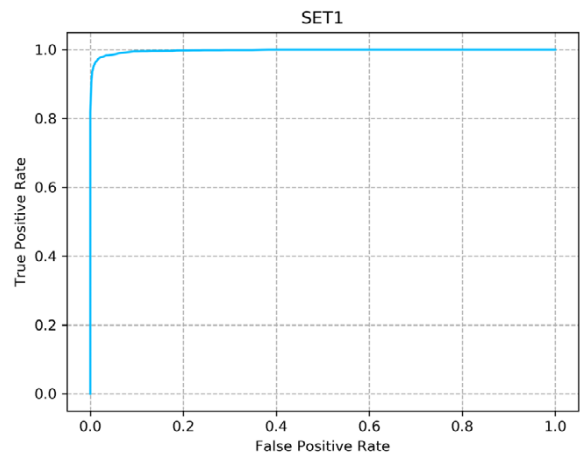

Figure 6. Receiver Operating Characteristic curve (ROC)

7674 individuals with each has around 90 pictures with different poses. We split the datasets into two disjoint sets, selecting 7098 people with 636252 images as training set, and 576 people with 51602 images as validation set. During the training, we generate the genuine and impostor pairs with a 50% probability, the same for validation. Additionally, the input images are cropped to the size 120×120.

The test dataset contains two parts, one part comes from the validation set, namely SET1, which is used to evaluate the robustness of 3D reconstruction and recognition. Another one is the AFLW2000-3D [18], namely SET2, which is used to evaluate the accuracy of 3D reconstruction.

*B. 3D Face Reconstruction*

In this section, we consider the robustness and accuracy of our 3DMM estimates on the task of 3D face reconstruction on SET1 and SET2, by comparing with 3DDFA. Following [16], we first reconstruct the 3D face shape and then globally align to the ground truth using standard, rigid iterative closest point (ICP) algorithm [20]. Then, we calculate the NME [5] normalized by the face bounding box size. Finally, we analyze NME's distribution of the same person with various poses and draw the boxplot in Fig. 4, we also show the quantitative evaluation results of reconstruction in Fig. 5. As can be seen from Fig. 4, our method is clearly more robustness than 3DDFA on SET1. The boxplots produced by SCNN have narrower widths than those by 3DDFA, indicating the reconstructed shapes by SCNN from different poses have smaller variations; Fig. 5 shows the comparison results between SCNN and 3DDFA on SET2, our SCNN achieves the NME of 35.98%, very close to the NME of 3DDFA (35.99%), indicating that our SCNN performs 3D face reconstruction as good as 3DDFA. However, our SCNN achieves much more robust results than 3DDFA, shown by Fig. 4.

*C. Face Recognition under Limited Conditions*

Since our target is improving robustness of 3D face reconstruction of the same individual under different poses, and meanwhile preserve the discriminative identity information in the feature space. Therefore, we do not try the best to set new face recognition records and test it in the wild. Following [3], we randomly selected 6000 pairs images from SET1, which contains 3000 genuine pairs and 3000 impostor pairs. We report performance on 10-fold cross validation using the split sub-datasets we have randomly generated. As can be seen from Fig. 6, our method achieves the recognition accuracy of 97.82%, indicating our model can extract discriminative facial feature while reconstruct robust 3D face shapes.

CONCLUSION

In this paper, we employ a novel Siamese Convolutional Neural Network (SCNN) method for robust 3D face reconstruction and discriminative identity information extraction. To this end, we introduce three losses to ensure the accuracy and robustness of face reconstruction, while preserving the identity information of input image. Experiments on two face datasets illustrate the effectiveness of SCNN on both 3D face reconstruction and 2D face recognition by comparing with other state-of-the-arts.